\documentclass[journal]{IEEEtran}
\usepackage[table]{xcolor}
\usepackage{cite}
\usepackage{amsmath,amssymb,amsfonts}
\usepackage{algorithm}
\usepackage{algorithmic}
\usepackage{graphicx}
\usepackage{subcaption}
\usepackage{textcomp}
\usepackage{array}
\usepackage{bm}
\usepackage{booktabs}
\usepackage{multirow}

\newcolumntype{P}[1]{>{\centering\arraybackslash}p{#1}}
\newcolumntype{M}[1]{>{\centering\arraybackslash}m{#1} }
\def\BibTeX{{\rm B\kern-.05em{\sc i\kern-.025em b}\kern-.08em
    T\kern-.1667em\lower.7ex\hbox{E}\kern-.125emX}}
\usepackage{tikz}
\usetikzlibrary{shapes,shapes.geometric,arrows,positioning,calc,decorations.pathmorphing,decorations.pathreplacing}

\begin{document}
\title{Inductive Power Grid Cascading Failure Analysis with GRU-Gated Graph Attention}

\author{\IEEEauthorblockN{Tianxin Zhou\IEEEauthorrefmark{1}, Xiang Li\IEEEauthorrefmark{1}, Haibing Lu\IEEEauthorrefmark{2}}\\
    \IEEEauthorblockA{\IEEEauthorrefmark{1}\textit{Department of Computer Science \& Engineering}, \\
    \IEEEauthorrefmark{2}\textit{Dept. of Information Systems \& Analytics} \\
    \textit{Santa Clara University}\\
    Santa Clara, USA \\
    \{tzhou,xli8,hlu\}@scu.edu}
    }

\maketitle

\begin{abstract}
Identifying vulnerable transmission lines in power grids before a cascading failure occurs is challenging: existing methods can learn inter-line failure correlations from cascade data, but they are trained and evaluated on a single grid, and transferring the learned knowledge to an unseen grid remains an open problem.
We address this by training a single Gated Recurrent Unit (GRU)-gated Graph Attention Network on combined cascading failure data from limited training grids and applying it directly to any unseen grid without retraining.
A GRU gate controls what information each node retains or discards at each cascade iteration.
Empirical evaluation shows that the model transfers zero-shot to multiple new grids spanning inter-time and inter-domain settings.
Using information extracted from the trained model, we consistently identify more vulnerable lines than established structural and electrical baselines.
\end{abstract}


\section{Introduction}\label{Introduction}
Cascading failure in power grids (PGCF) remains one of the primary causes of large-scale blackouts worldwide~\cite{NE03report,Italy03report}.
A single overloaded transmission line can trigger a cascading failure that, in the worst case, collapses an entire power network.
Identifying which lines are most vulnerable to such failures is therefore a practical priority for power grid security.

Existing approaches to PGCF vulnerability analysis range from deterministic $N$-$k$ contingency methods~\cite{pg14gil,nk10bienstock,opa02dobson} to stochastic models~\cite{probabilistic14gupta,markov20nakarmi}, supervised machine learning~\cite{svm15gupta,bayes18pi,ml19shuvro}, self-supervised attention methods~\cite{dual25zhou}, and graph neural networks (GNN)~\cite{nn21liao,severity24gorka,prediction24bhaila,prediction23chadaga}.
A key observation from~\cite{pg14gil} is that line failure is statistically independent of topological distance to the preceding failure (Lemma~4.1), which makes feature-based prediction ineffective and motivates learning propagation patterns directly from cascade data.

Vulnerability identification requires knowing which lines are most implicated across the full space of possible cascades.
Any simulation-based approach must enumerate $\binom{N}{k}$ initial failure combinations to answer this question --- intractable regardless of how fast the underlying power-flow computation is~\cite{nk10bienstock}.
The self-supervised attention method of~\cite{dual25zhou} sidesteps this enumeration by learning line-level vulnerability directly from cascade observations, but it trains a separate model per grid and cannot transfer to an unseen grid.
GNN methods for cascading failure~\cite{risk23zhu,cf22zhu,prediction23chadaga} are likewise confined to a single grid; moreover, they are not designed for vulnerability identification at all --- they output system-level simulation metrics or cascade predictions --- and if repurposed for it, they would face the same $\binom{N}{k}$ enumeration wall.
Cross-grid GNN methods~\cite{gnn25zhu,pf23hansen} improve generalization for other power system tasks such as power flow approximation and contingency screening, but they also do not target vulnerability identification and require supervised fine-tuning on labeled data from the evaluation grid before deployment.
To our knowledge, no existing work identifies line-level vulnerability zero-shot on a structurally unseen grid. This gap is practically significant because grid topologies evolve as lines are added, retired, or rerouted. Existing methods would therefore need fine-tuning or retraining for each new topology, including solving power-flow equations to regenerate labeled data.

We address this gap through \textit{cross-grid learning}: training a single model on cascade data from a small set of grids and applying it directly to any unseen grid without retraining.
Our model, \textbf{CG-CAE} (Cross-Grid Cascade Attention Exposure), pairs a Gated Recurrent Unit (GRU)-gated graph attention architecture with a cascade-depth mask that filters attention to causally relevant failures.
The filtered attention coefficients are aggregated into a per-line cascade exposure score that ranks lines by their involvement in failure propagation.
We make the following contributions.

\begin{enumerate}

    \item We propose a GRU-gated graph attention architecture that propagates cascade state across iterations recurrently.
    The vector update gate~\cite{gru14cho} allows each node to relearn, reset, or blend information from previous propagation steps in a data-driven manner.
    The architecture is fully inductive and transfers zero-shot to new grids without fine-tuning, eliminating the need to solve power-flow equations on the evaluation grid to generate supervised fine-tuning data.

    \item We introduce a cascade-aware attention aggregation that employs a \textit{cascade-depth} mask restricting the vulnerability signal to lines that are causally subsequent to the initial failures in the cascade.
    The filtered attention coefficients yield a \textit{cascade exposure} score for each line, enabling operators to identify the most vulnerable lines.

    \item We demonstrate that a single model trained on limited grids transfers zero-shot to new grids with no fine-tuning or retraining.
    This cross-grid generalization relies on shared-parameter message-passing, a property absent from prior per-grid approaches such as~\cite{dual25zhou}.
    The cascade exposure score identifies more vulnerable lines than Electric Betweenness and PageRank baselines at every tested top-$\tau\%$ threshold.

\end{enumerate}

\textbf{Organization.}
Section~\ref{RelatedWork} reviews related work on cascading failure analysis and GNN applications.
Section~\ref{CAE} presents the CG-CAE model.
Section~\ref{Experiments} describes the experimental setup and discusses the results.
We conclude in Section~\ref{Conclusion}.

\section{Related Work}\label{RelatedWork}

Methods for vulnerability analysis against PGCF can be broadly grouped into five directions.
1)~Deterministic methods compute power-flow solutions directly to evaluate $N$-$k$ contingencies~\cite{pg14gil,nk10bienstock,opa02dobson}.
While physically accurate, they are computationally intractable for large networks.
2)~Stochastic methods model line-failure probabilities via distributions or Markov processes~\cite{probabilistic14gupta,markov20nakarmi}.
3)~Conventional machine learning treats PGCF prediction as a supervised problem using physical and topological features as input~\cite{svm15gupta,bayes18pi,ml19shuvro}.
Both stochastic and conventional machine-learning methods operate at the system level --- predicting cascade occurrence, cascade size, or overall grid resilience --- rather than at the component level.
Their outputs can broadly categorize risk but cannot pinpoint which individual lines are vulnerable or the order in which they fail.

4)~Self-supervised attention-based methods learn inter-line failure correlations directly from cascade samples~\cite{dual25zhou}.
This is significant because the failure of a line is statistically independent of its topological distance to the preceding failure (Lemma~4.1,~\cite{pg14gil}), a property that makes conventional label-driven supervision ineffective at capturing propagation patterns.
By training a transformer-based model in a self-supervised manner, \cite{dual25zhou} recovers these hidden correlations implicitly.
However, \cite{dual25zhou} trains a separate model per grid, which limits its applicability when the grid topology changes.

5)~GNN methods exploit the graph structure of the power network, and their adoption in power system analysis has grown rapidly~\cite{nn21liao,severity24gorka,prediction24bhaila,prediction23chadaga}.
For cascading failure, \cite{risk23zhu} uses a graph convolutional network to evaluate real-time cascading failure risk under high renewable penetration.
The method in~\cite{cf22zhu} replaces the power-flow computation inside a cascading failure simulator with a physics-informed GNN, accelerating simulation while generalizing to topology perturbations within a single base grid via pre-training and fine-tuning; however, its output is system-level simulation metrics (branch-loss and load-loss ratios), not line-level vulnerability rankings, and the trained model cannot be applied to a structurally different grid.
Chadaga et al.~\cite{prediction23chadaga} proposes a flow-free GNN that predicts the full cascade sequence at each iteration without intermediate power-flow computation.
All three methods train and evaluate on a single grid and are designed for simulation or prediction tasks, not for vulnerability identification.
Beyond cascading failure, self-supervised pre-training has been applied to GNNs for broader power system tasks.
Zhu et al.~\cite{gnn25zhu} pre-train a Graph Transformer with physics-informed masked-feature and masked-edge prediction, building on the self-supervised graph learning framework of~\cite{ssl20rong}, and fine-tune on downstream tasks --- power flow, optimal power flow, contingency screening, and outage prediction --- each requiring labeled data generated by solving power-flow equations.
Hansen et al.~\cite{pf23hansen} trains a fully decentralized GNN simultaneously on six IEEE grids for power flow balancing and evaluates on two structurally unseen grids, showing that multi-topology training improves cross-grid generalization.
Both methods share self-supervised or multi-grid pre-training and inductive architectures with our approach; however, they address prediction and regression tasks rather than vulnerability identification, and both require supervised fine-tuning on the evaluation grid --- neither achieves zero-shot transfer.

In summary, existing GNN methods for cascading failure~\cite{risk23zhu,cf22zhu,prediction23chadaga} are confined to a single grid and produce simulation metrics or cascade predictions, not vulnerability rankings.
Cross-grid GNN methods~\cite{gnn25zhu,pf23hansen} generalize across topologies but solve prediction tasks and require supervised fine-tuning with labeled data from each new grid.
The self-supervised attention method of~\cite{dual25zhou} directly targets line-level vulnerability identification by learning from cascade data, avoiding the $\binom{N}{k}$ enumeration that simulation-based approaches would face, but it is limited to a single grid.
CG-CAE extends this line of work to the cross-grid setting: it learns vulnerability from cascade observations and transfers zero-shot to structurally unseen grids without fine-tuning or solving power-flow equations on the evaluation grid.

\section{CG-CAE Model}\label{CAE}
We propose CG-CAE (Cross-Grid Cascade Attention Exposure) for inductive power grid cascading failure analysis.
The backbone of CG-CAE is the GRU-gated graph attention architecture shown in Fig.~\ref{fig:grugat}.
A Graph Attention Network (GAT) layer provides message-passing whose attention coefficients yield per-edge signals for cascade exposure extraction, while a GRU gate controls what each node retains or discards at each iteration to handle variable cascade depth.
A cascade-depth mask then filters attention to causally informed source nodes, and the filtered coefficients are aggregated into a per-line cascade exposure score.
The remainder of this section is organized as follows: we introduce the cascading failure model (\ref{CAE:CF}), define the line graph (\ref{CAE:LG}), describe the GRU-gated graph attention architecture (\ref{CAE:GRU}), present the cascade-aware attention analysis that produces vulnerability rankings (\ref{CAE:attention}), and summarize the cross-grid training and deployment framework (\ref{CAE:framework}).

\subsection{Cascading Failure Model}\label{CAE:CF}

\begin{figure}[ht]
    \centering
    \begin{tikzpicture}[
        x=\columnwidth/40,
        y=\columnwidth/40,
        node distance=2,
        every node/.style={font=\normalsize},
        block/.style={
            rectangle,
            draw,
            text centered,
            rounded corners=3,
        },
        decision/.style={
            diamond,
            draw,
            text centered,
            aspect=1.5,
        },
        terminal/.style={
            rounded rectangle,
            draw,
            text centered,
            rounded corners=5,
        },
        arrow/.style={-latex, thick, line width=0.5}
    ]

    \node[terminal, text width=30mm] (init) {Initialize $S_1$, $g=1$};
    \node[block, below=of init, text width=30mm] (disconnect) {Disconnect failed lines in $S_{g-1}$};
    \node[block, below=of disconnect, text width=30mm] (findf) {Recompute flows, update $S_g$};
    \node[decision, below=of findf] (decide1) {$S_{g} = S_{g-1}$?};
    \node[terminal, below=of decide1, text width=30mm] (output) {Output $S_{g}$};
    \node[block, right=12mm of findf, text width=28mm] (update) {$g = g + 1$};

    \draw[arrow] (init) -- (disconnect);
    \draw[arrow] (disconnect) -- (findf);
    \draw[arrow] (findf) -- (decide1);
    \draw[arrow] (decide1) -- node[right] {Yes} (output);
    \draw[arrow] (decide1) -- node[above] {No} (update);
    \draw[arrow] (update) -| ($(disconnect.east)+(5mm,0)$) -- (disconnect);
    \end{tikzpicture}

    \caption{Cascading Failure Model.}
    \label{fig:cf}
\end{figure}
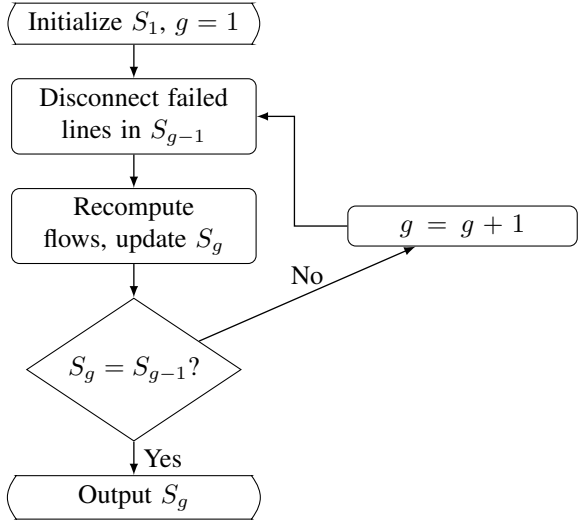

The cascading failure process, as described in~\cite{pg14gil} and illustrated in Fig.~\ref{fig:cf}, begins with all lines in an initial state $S_{1}$: each line is either operational ($0$) or failed ($1$).
Some lines are designated as initial failures at iteration $g=1$.
At each iteration $g$, the state $S_{g}$ of all lines is updated: failed lines are disconnected, and the power flows on the remaining operational lines are recomputed via the power-flow equations in~\cite{pg14gil}.
A new failure occurs when the power flow $f_{ij}$ on an operational line exceeds its capacity, updating the state according to
\begin{equation}\label{eq:state}
s_{ij} = \begin{cases}
    0 & \text{operational} \\
    g & \text{failed at iteration } g
\end{cases}
\end{equation}
for each line $(i,j)$.
The process iterates until $S_{g} = S_{g-1}$, at which point the cascade has converged.
Samples that converge at $g = 1$ did not propagate beyond the initial failures and are discarded, as they carry no propagation signal for training.

This simulator produces cascade samples grounded in DC power-flow equations, providing the data CG-CAE uses for training and evaluation.

\subsection{Line Graph Construction}\label{CAE:LG}
In the original power network, buses are nodes and transmission lines are edges, but  cascading failure analysis requires reasoning about the lines themselves.

Line graph representations have been applied in power system GNN work for static security analysis~\cite{switching23ye}; we adopt the same transformation because it lets the model operate directly on line-to-line relationships, the natural granularity for  vulnerability identification.

\begin{figure}[ht]
    \centering
    \begin{tikzpicture}[scale=1, every node/.style={font=\small}]
        \node[circle, draw, fill=blue!10] (A) at (0,1.5) {$B_i$};
        \node[circle, draw, fill=blue!10] (B) at (2,1.5) {$B_j$};
        \node[circle, draw, fill=blue!10] (C) at (1,0) {$B_k$};
        \draw[thick] (A) -- node[above] {$L_u$} (B);
        \draw[thick] (B) -- node[right] {$L_v$} (C);
        \draw[thick] (C) -- node[left] {$L_w$} (A);
        \node at (1,-1) {Original Power Network};

        \node[rectangle, draw, fill=red!10] (E1) at (5,1.5) {$L_u$};
        \node[rectangle, draw, fill=red!10] (E2) at (7,1.5) {$L_v$};
        \node[rectangle, draw, fill=red!10] (E3) at (6,0) {$L_w$};
        \draw[thick] (E1) -- node[above] {$B_j$} (E2);
        \draw[thick] (E2) -- node[right] {$B_k$} (E3);
        \draw[thick] (E3) -- node[left] {$B_i$} (E1);

        \node at (6,-1) {Line graph};
    \end{tikzpicture}
    \caption{Transformation from power network to line graph: transmission lines become nodes, and shared buses induce edges.}
    \label{fig:linegraph}
\end{figure}
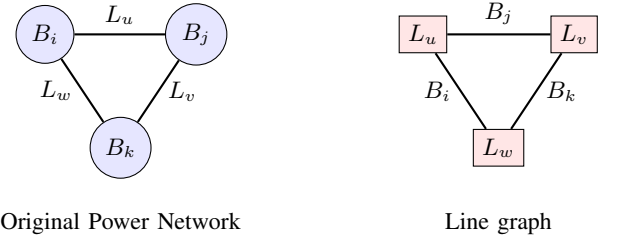

Hence, we convert the network into a \emph{line graph}: each transmission line becomes a node, and an edge is drawn between two nodes if the corresponding lines share a common bus.
This transformation is illustrated in Fig.~\ref{fig:linegraph}.

Concretely, each line graph node $u$ corresponds to a transmission line $L_u$ connecting buses $(B_i, B_j)$.
An edge $(u, v)$ exists if $L_u$ and $L_v$ share a bus, e.g.\ both connect to $B_j$.

\subsection{GRU-Gated Graph Attention Architecture}\label{CAE:GRU}
CG-CAE processes each cascade sample through three stages: a failure-iteration embedding maps each line's ordinal depth to a learned vector (\ref{CAE:Embed}), a GAT layer performs message-passing over the line graph (\ref{CAE:GAT}), and a GRU gate controls what each node retains across cascade iterations (\ref{CAE:GRU:recurrent}).
\subsubsection{Failure Iteration Embedding}\label{CAE:Embed}
Each node's input feature is the failure iteration of its corresponding transmission line:
\begin{equation}\label{eq:nf}
    \mathbf{h}_u = \mathbf{E}[g_u]
\end{equation}
where $\mathbf{E}$ is a trainable lookup table and $g_u$ is the failure iteration of line $u$ (0 if it did not fail).
Because cascade modeling depends on \emph{when} a line failed rather than which specific line it is, this ordinal encoding captures the essential temporal information.
Early-iteration failures naturally carry different information than late-iteration ones, and the learned embedding distinguishes these patterns from the outset.
Moreover, because the representation depends only on an ordinal iteration index and not on any grid-specific node identifier, the same embedding transfers directly to any grid.
While physical line attributes such as load and capacity could supplement the input, they require grid-specific normalization and may not be consistently available across grids, complicating zero-shot deployment.
The ordinal embedding requires no external data beyond the cascade sample itself, keeping the input self-contained and grid-agnostic.

\subsubsection{Graph Attention Layer}\label{CAE:GAT}

Our model builds on GATs~\cite{gat18velickovic}, which assign learned weights to messages between connected nodes.
For each directed edge $(u, v)$, GAT computes an attention coefficient
\begin{equation}\label{eq:ea}
    \alpha_{uv} = \frac{
        \exp\!\left(\text{LeakyReLU}\!\left(\mathbf{a}^T[\mathbf{W}\mathbf{h}_u\,\|\,\mathbf{W}\mathbf{h}_v]\right)\right)
    }{
        \sum_{w\in\mathcal{N}(u)}\exp\!\left(\text{LeakyReLU}\!\left(\mathbf{a}^T[\mathbf{W}\mathbf{h}_u\,\|\,\mathbf{W}\mathbf{h}_w]\right)\right)
    }
\end{equation}
 where $\mathbf{a}$ is a trainable weight vector, $\mathbf{W}$ is a shared linear transformation, and $\mathbf{h}_u$ is the current feature vector of node $u$.
 The updated node representation aggregates over its neighborhood:
\begin{equation}\label{eq:vr}
    \mathbf{h}'_u = \sigma\!\left(\sum_{v\in\mathcal{N}(u)} \alpha_{uv}\,\mathbf{W}\mathbf{h}_v\right)
\end{equation}
 where $\sigma$ is a nonlinear activation.
 Because $\mathbf{W}$ and $\mathbf{a}$ are shared across the entire graph, the same parameters apply to any topology.
 This shared-parameter design, common to message-passing neural networks, is what makes CG-CAE inductive.
 We choose GAT over other message-passing architectures because its attention coefficients $\alpha_{uv}$ provide per-edge signals that we later filter and aggregate into the cascade exposure score.

\subsubsection{GRU-Gated Recurrent Propagation}\label{CAE:GRU:recurrent}
A standard  GAT applies a fixed number of message-passing rounds regardless of cascade depth.
This is limiting for two reasons: the number of iterations $G$ varies between samples, and the information a node should carry forward evolves as the cascade progresses.
Early-iteration failure locations should remain influential throughout, yet later iterations introduce new failure messages that only some nodes need to integrate.
A GRU gate applied after each cascade iteration addresses both issues.

\begin{figure}[ht]
    \centering
    \begin{tikzpicture}[
        every node/.style={font=\small},
        ebox/.style  ={rectangle, draw, rounded corners=2pt, fill=green!15,
                       minimum height=7mm, minimum width=32mm, align=center},
        ln/.style    ={rectangle, draw, rounded corners=2pt, fill=blue!8,
                       minimum height=7mm, minimum width=32mm, align=center},
        gat/.style   ={rectangle, draw, rounded corners=2pt, fill=blue!15,
                       minimum height=7mm, minimum width=32mm, align=center},
        gru/.style   ={rectangle, draw, rounded corners=2pt, fill=orange!18,
                   minimum height=7mm, minimum width=32mm, align=center},
        outbox/.style={rectangle, draw, rounded corners=2pt, fill=red!15,
                       minimum height=7mm, minimum width=32mm, align=center},
        arr/.style={-latex, thick}
    ]
    \node[ebox] (emb) at (0,0) {Embedding $\mathbf{E}[g_u]$};

    \node[ln, below=8mm of emb] (ln1) {LayerNorm};
    \node[gat, below=4mm of ln1] (gat) {GAT};
    \node[gru, below=4mm of gat] (gru) {GRU Gate};
    \draw[dashed, gray!70, rounded corners=5pt]
        ($(ln1.north west)+(-6mm,+1.5mm)$) rectangle ($(gru.south east)+(+6mm,-1.5mm)$);

    \node[font=\scriptsize, text=gray!70, align=left]
          at ($(ln1.north west)+(3mm,+3mm)$)
          {shared weights};

    \node[ln, below=10mm of gru] (ln2) {LayerNorm};
    \node[outbox, below=4mm of ln2] (outn) {GAT$_{\text{out}}$};
    \node[below=4mm of outn] (logits) {logits};
    \draw[arr] (emb.south)
        -- node[right, font=\small]{$\mathbf{h}^{(0)}$}
        (ln1.north);

    \draw[arr] (ln1.south) -- (gat.north);
    \draw[arr] (gat.south) -- (gru.north);
    \draw[arr] (gat.west)
        -- ++(-16mm,0)
        node[left, font=\small]{$\alpha^{(k,t)}_{uv}$};

    \coordinate (hfork1) at ($(gru.south)+(0,-4mm)$);
    \fill (hfork1) circle (0.9pt);

    \coordinate (hfork2) at ($(gru.east)+(16mm,0)$);
    \fill (hfork2) circle (0.9pt);

    \draw[thick] (gru.south) -- (hfork1);
    \draw[arr] (hfork1)
        -- node[right, font=\small]{$\mathbf{h}^{(G_{(k)}-1)}$}
        (ln2.north);

    \draw[thick] (hfork1) -- ($(hfork1 -| hfork2)$) -- (hfork2);
    \draw[arr] (hfork2)
        -- (gru.east);

    \draw[arr] (hfork2)
        node[right, font=\small]{$\mathbf{h}_{\text{new}}^{(t)}$}
        -- ($(hfork2 |- ln1.east)$)
        -- (ln1.east);

    \draw[arr] (ln2.south) -- (outn.north);
    \draw[arr] (outn.south) -- (logits.north);

    \end{tikzpicture}
    \caption{GRU-GAT recurrent architecture.
             Arrows trace the data flow: the failure-iteration embedding produces $\mathbf{h}^{(0)}$, which enters the shared-weight loop (dashed box).
             The GRU gate (Eqs.~\eqref{eq:gru_z}--\eqref{eq:gru_out}) decides how much of the previous hidden state $\mathbf{h}^{(t-1)}$ to retain versus how much of the new GAT message $\mathbf{h}^{(t)}_{\text{new}}$ to accept, producing $\mathbf{h}^{(t)}$.
             The feedback path (right) carries $\mathbf{h}^{(t)}$ back into the loop.
             The GAT layer emits attention coefficients $\alpha^{(k,t)}_{uv}$ (left) to be filtered and aggregated in Sec.~\ref{ATT:agg} (Eq.~\eqref{eq:att_in}).
             The final LayerNorm and output GAT layer yield per-node logits for cross-grid training.}
    \label{fig:grugat}
\end{figure}
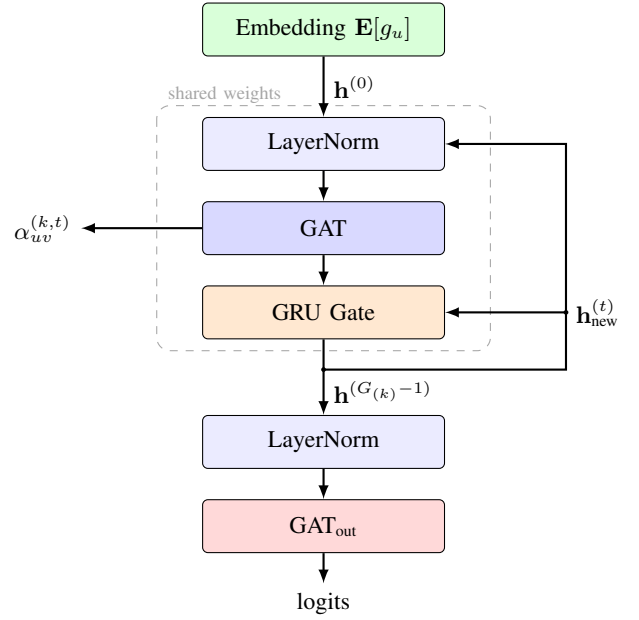

The structure of the GRU-GAT loop is illustrated in Fig.~\ref{fig:grugat}.
For a cascade sample $k$ with maximum failure iteration $G^{(k)}$, we run $G^{(k)} - 1$ GRU-GAT steps over the \emph{hidden} layers, followed by a single output layer.
Each node starts from its failure-iteration embedding $\mathbf{h}^{(0)} = \mathbf{E}[g_u]$.
At each GRU-GAT step $t$, the GAT layer produces a new message:

\label{CAE:GRU:GAT}
\begin{equation}\label{eq:message_passing}
    \mathbf{h}^{(t)}_{\text{new}} = \text{ELU}\!\left(
         \text{GAT}\!\left(\text{LN}(\mathbf{h}^{(t-1)})\right)
    \right)
\end{equation}
LayerNorm stabilizes the input at each step.
The GRU gate then decides how much of the previous state to retain:
\label{CAE:GRU:Gate}
\begin{align}
    \mathbf{z}^{(t)} &= \sigma\!\left(
        \mathbf{W}_z\!\left[\mathbf{h}^{(t-1)} \,\|\, \mathbf{h}^{(t)}_{\text{new}}\right]
    \right) \in \mathbb{R}^{N \times H} \label{eq:gru_z} \\
    \mathbf{r}^{(t)} &= \sigma\!\left(
        \mathbf{W}_r\!\left[\mathbf{h}^{(t-1)} \,\|\, \mathbf{h}^{(t)}_{\text{new}}\right]
    \right) \label{eq:gru_r} \\
    \tilde{\mathbf{h}}^{(t)} &= \tanh\!\left(
        \mathbf{W}_h\!\left[\mathbf{r}^{(t)} \odot \mathbf{h}^{(t-1)} \,\|\, \mathbf{h}^{(t)}_{\text{new}}\right]
    \right) \label{eq:gru_h} \\
    \mathbf{h}^{(t)} &= (1 - \mathbf{z}^{(t)}) \odot \mathbf{h}^{(t-1)} +
                        \mathbf{z}^{(t)} \odot \tilde{\mathbf{h}}^{(t)} \label{eq:gru_out}
\end{align}
These four operations constitute the ``GRU Gate'' cell shown in Fig.~\ref{fig:grugat}: it receives the previous hidden state $\mathbf{h}^{(t-1)}$ and the GAT message $\mathbf{h}^{(t)}_{\text{new}}$, and produces the updated state $\mathbf{h}^{(t)}$ that is fed back into the next step of the shared-weight loop.

\label{CAE:GRU:Output}
After the final GRU-GAT step, a second LayerNorm and an output GAT layer produce per-node class logits:
\begin{equation}\label{eq:output}
    \text{logits} =  \text{GAT}_{\text{out}}\!\left(\text{LN}(\mathbf{h}^{(G_{(k)}-1)})\right)
\end{equation}

The same $(\mathbf{W}_z, \mathbf{W}_r, \mathbf{W}_h)$ and  GAT weights are reused at every GRU-GAT step, keeping the parameter count constant regardless of cascade depth and forcing the model to learn a general update rule.

We run exactly $G_{(k)} - 1$ steps for each sample $k$.
Each message-passing step corresponds to one iteration of the cascading process, so the model propagates information along the cascade progression.

Training is self-supervised: given a complete cascade sample with all failure iterations known, the model learns to reconstruct the cascade pattern.
 This objective encourages the model to capture how failures at different iterations relate to one another.
 The attention weights and hidden states encode these inter-line failure correlations, which we later extract for cascade exposure analysis.

\subsection{Cascade-Aware Attention Analysis}\label{CAE:attention}
Once training is complete, the failure correlations captured by the model must be translated into actionable vulnerability information.
The attention matrix is a natural source for this: when the model consistently directs high attention toward a node $v$ across many cascade scenarios, it signals that failures in neighboring lines are correlated with $v$'s involvement in cascades.
The challenge is that not all attention coefficients are equally informative.

Below, we describe how attention is filtered and aggregated into a cascade exposure score that identifies the most vulnerable transmission lines.
Fig.~\ref{fig:cae} summarizes the complete pipeline from raw attention to final ranking.

\subsubsection{Cascade-Aware Attention Masking}\label{ATT:mask}
At GRU-GAT step $t$, the GAT layer in Eq.~\eqref{eq:message_passing} computes attention coefficients $\alpha_{uv}$ for every directed edge $(u, v)$ in the line graph.
Here, $u$ is the source node that sends a message and $v$ is the target node that receives it --- the attention coefficient $\alpha_{uv}$ measures how much the model weights the message from $u$ when updating $v$'s representation.
However, only some of those edges carry genuine cascade-propagation messages at step $t$.
Specifically, a source node $u$ has a meaningful representation at step $t$ only if its cascade depth satisfies $d_u \leq t$, where $d_u$ is the shortest-path distance in the line graph from node $u$ to the nearest initial failure node.
Nodes with greater cascade depth have not yet received cascade context through message-passing and their attention is largely noise.

Therefore, we apply the mask
\begin{equation}\label{eq:att_mask}
    \mathcal{M}^{(t)} = \left\{(u,v) \;\middle|\; d_u \leq t \right\}
\end{equation}
This mask ensures that we only collect attention from source nodes whose representations have been informed by at least $t$ rounds of cascade context.

As shown in Fig.~\ref{fig:grugat}, we collect attention only from the hidden GRU-GAT steps.
The output GAT layer (Eq.~\eqref{eq:output}) serves the classification objective and is excluded from attention aggregation.

\subsubsection{Cascade-Size-Weighted Aggregation}\label{ATT:agg}
For each cascade sample $k$, we compute the masked incoming attention --- the total attention directed toward each target node $v$ from source nodes that pass the cascade-depth mask:
\begin{equation}\label{eq:att_in}
    a^{(k)}_v = \sum_{t=0}^{G^{(k)}-2} \sum_{(u,v) \in \mathcal{M}^{(t)}} \alpha^{(k,t)}_{uv}
\end{equation}
where $\alpha^{(k,t)}_{uv}$ is the head-averaged attention coefficient at step $t$ in sample $k$, and $G^{(k)}$ is the maximum failure iteration in that sample.
The upper limit is $G^{(k)} - 2$ because the model runs $G^{(k)} - 1$ hidden steps indexed from $t = 0$ to $t = G^{(k)} - 2$; the output layer at step $G^{(k)} - 1$ is excluded from attention collection.

We then weight each sample's contribution by its cascade size, defined as the number of lines that failed beyond the initial iteration:
\begin{equation}\label{eq:cascade_weight}
    w_k = \left|\left\{u : g^{(k)}_u > 1\right\}\right|
\end{equation}
Larger cascades are more informative about propagation patterns, so they receive greater weight.
The aggregated attention score for each node is
\begin{equation}\label{eq:agg}
    A_v = \frac{\sum_k w_k \cdot a^{(k)}_v}{\sum_k w_k}
\end{equation}

\begin{figure*}[ht]
    \centering
    \begin{tikzpicture}[
        every node/.style={font=\small},
        lginf/.style={rectangle, draw, thick, fill=red!10, minimum size=7.5mm, font=\small},
        lguninf/.style={rectangle, draw, thick, fill=gray!15, minimum size=7.5mm, font=\small, text=gray!55},
        live/.style={-latex, thick, blue!70},
        dead/.style={-latex, thin, dashed, gray!40},
        dlabel/.style={font=\scriptsize, text=black!70},
        outerbox/.style={rectangle, draw, rounded corners=2pt, fill=blue!10,
                         minimum height=6mm, minimum width=10mm, align=center, font=\tiny},
        innerbox/.style={rectangle, draw, rounded corners=2pt, fill=orange!15,
                         minimum height=6mm, minimum width=8mm, align=center, font=\scriptsize},
        centerbox/.style={circle, draw, thick, fill=green!20,
                          minimum size=8mm, align=center, font=\normalsize},
        wtlabel/.style={font=\scriptsize, text=black!50},
        ringarr/.style={-latex, thin, gray!50},
        dotnode/.style={font=\normalsize, text=gray!50},
        eqnote/.style={font=\scriptsize, text=black!55},
    ]

    \node[font=\small\bfseries] at (1.1, 3.2) {(a) Cascade-Depth Mask};

    \node[font=\small\bfseries] at (1.1, 2.5) {$t=0$};

    \node[lginf]   (u0) at (0.0, 1.2) {$L_u$};
    \node[lguninf] (v0) at (2.2, 1.2) {$L_v$};
    \node[lguninf] (w0) at (1.1,-0.7) {$L_w$};

    \node[dlabel, above=2pt of u0] {$d_u{=}0$};
    \node[dlabel, above=2pt of v0] {$d_v{=}1$};
    \node[dlabel, below=2pt of w0] {$d_w{=}1$};

    \draw[live] ([yshift=1mm]u0.east) -- node[above, font=\scriptsize, text=blue!70] {$\alpha_{uv}$} ([yshift=1mm]v0.west);
    \draw[live] ([xshift=-2mm]u0.south) -- node[left, font=\scriptsize, text=blue!70, pos=0.35] {$\alpha_{uw}$} ([xshift=-3mm]w0.north);

    \draw[dead] ([yshift=-1mm]v0.west) -- ([yshift=-1mm]u0.east);
    \draw[dead] ([xshift=2mm]v0.south) -- ([xshift=3mm]w0.north);
    \draw[dead] ([xshift=1mm]w0.north) -- ([xshift=2mm]u0.south);
    \draw[dead] ([xshift=-1mm]w0.north east) -- ([xshift=-2mm]v0.south);

    \node[font=\scriptsize, text=blue!70, align=center] at (1.1, -1.8)
          {$\mathcal{M}^{(0)}\!:$ only $L_u$ emits};

    \node[font=\small\bfseries] at (6.3, 2.5) {$t=1$};

    \node[lginf] (u1) at (5.0, 1.2) {$L_u$};
    \node[lginf] (v1) at (7.6, 1.2) {$L_v$};
    \node[lginf] (w1) at (6.3,-0.7) {$L_w$};

    \node[dlabel, above=2pt of u1] {$d_u{=}0$};
    \node[dlabel, above=2pt of v1] {$d_v{=}1$};
    \node[dlabel, below=2pt of w1] {$d_w{=}1$};

    \draw[live] ([yshift=1.5mm]u1.east) -- node[above, font=\scriptsize, text=blue!70] {$\alpha_{uv}$} ([yshift=1.5mm]v1.west);
    \draw[live] ([yshift=-1.5mm]v1.west) -- node[below, font=\scriptsize, text=blue!70] {$\alpha_{vu}$} ([yshift=-1.5mm]u1.east);

    \draw[live] ([xshift=-2mm]u1.south) -- node[left, font=\scriptsize, text=blue!70, pos=0.32] {$\alpha_{uw}$} ([xshift=-4mm]w1.north);
    \draw[live] ([xshift=-1mm]w1.north) -- node[right, font=\scriptsize, text=blue!70, pos=0.65] {$\alpha_{wu}$} ([xshift=0mm]u1.south);

    \draw[live] ([xshift=2mm]v1.south) -- node[right, font=\scriptsize, text=blue!70, pos=0.32] {$\alpha_{vw}$} ([xshift=4mm]w1.north);
    \draw[live] ([xshift=1mm]w1.north) -- node[left, font=\scriptsize, text=blue!70, pos=0.65] {$\alpha_{wv}$} ([xshift=0mm]v1.south);

    \node[font=\scriptsize, text=blue!70, align=center] at (6.3, -1.8)
          {$\mathcal{M}^{(1)}\!:$ all nodes emit};

    \node[centerbox] (Av) at (13.0, 0.5) {$A_v$};

    \node[font=\small\bfseries, anchor=south west]
         at ([xshift=-3.5cm, yshift=2.7cm]Av.center) {(b) Aggregation for node $v$};

    \node[innerbox, above=6mm of Av]  (ak1) {$a^{(1)}_v$};
    \node[innerbox, right=6mm of Av]  (ak2) {$a^{(2)}_v$};
    \node[innerbox, below=6mm of Av]  (ak3) {$a^{(3)}_v$};
    \node[innerbox, left=6mm of Av]   (akK) {$a^{(K)}_v$};

    \draw[ringarr] (ak1) -- node[right, wtlabel, pos=0.45] {$w_1$} (Av);
    \draw[ringarr] (ak2) -- node[above, wtlabel, pos=0.45] {$w_2$} (Av);
    \draw[ringarr] (ak3) -- node[right, wtlabel, pos=0.45] {$w_3$} (Av);
    \draw[ringarr] (akK) -- node[above, wtlabel, pos=0.45] {$w_K$} (Av);

    \node[dotnode, below left=6mm and 6mm of Av] {$\cdots$};

    \node[outerbox, above left=2mm and 2mm of ak1]   (a1a) {$\alpha^{(1,0)}_{uv}$};
    \node[outerbox, above right=2mm and 2mm of ak1]  (a1b) {$\alpha^{(1,1)}_{wv}$};
    \draw[ringarr] (a1a) -- (ak1);
    \draw[ringarr] (a1b) -- (ak1);

    \node[outerbox, above right=2mm and 2mm of ak2]  (a2a) {$\alpha^{(2,0)}_{uv}$};
    \node[outerbox, below right=2mm and 2mm of ak2]  (a2b) {$\alpha^{(2,1)}_{wv}$};
    \draw[ringarr] (a2a) -- (ak2);
    \draw[ringarr] (a2b) -- (ak2);

    \node[outerbox, below left=2mm and 2mm of ak3]   (a3a) {$\alpha^{(3,0)}_{uv}$};
    \node[outerbox, below right=2mm and 2mm of ak3]  (a3b) {$\alpha^{(3,1)}_{wv}$};
    \draw[ringarr] (a3a) -- (ak3);
    \draw[ringarr] (a3b) -- (ak3);

    \node[outerbox, left=2mm of akK]  (aKa) {$\cdots$};
    \draw[ringarr] (aKa) -- (akK);

    \end{tikzpicture}
    \caption{Cascade-aware attention masking and aggregation.
             (a)~On the three-node line graph from Fig.~\ref{fig:linegraph}, red-shaded nodes pass the mask $\mathcal{M}^{(t)}$ ($d_u \leq t$) and emit attention (solid arrows); gray nodes are filtered out (dashed).
             At $t{=}0$ only the initial failure $L_u$ contributes; at $t{=}1$ all nodes satisfy $d \leq 1$.
             (b)~For a given node $v$, the filtered coefficients from each cascade sample $k$ are summed into $a^{(k)}_v$ (Eq.~\eqref{eq:att_in}, inner ring), then averaged with cascade-size weights $w_k = |\{u: g^{(k)}_u > 1\}|$ (Eq.~\eqref{eq:cascade_weight}) to produce the cascade exposure score $A_v$ (Eq.~\eqref{eq:agg}, center).}
    \label{fig:mask}\label{fig:aggregation}
\end{figure*}

\subsubsection{Cascade Exposure Score}\label{ATT:rank}
We define the \emph{cascade exposure} of a transmission line as its aggregated incoming attention score $A_v$ from Eq.~\eqref{eq:agg}.
All lines are ranked by $A_v$ in descending order: a higher score indicates that the model has learned stronger failure correlations between that line and other lines across many cascade scenarios.
Lines at the top of this ranking are the ones most exposed to cascading failure propagation.

\subsection{Cross-Grid Training and Framework}\label{CAE:framework}
Every learnable operation in  CG-CAE --- GAT attention (Eq.~\eqref{eq:ea}--\eqref{eq:vr}), GRU gate (Eq.~\eqref{eq:gru_z}--\eqref{eq:gru_out}), and output projection (Eq.~\eqref{eq:output}) --- shares a single parameter set $(\mathbf{W}, \mathbf{a}, \mathbf{W}_z, \mathbf{W}_r, \mathbf{W}_h)$.
Because no component depends on a fixed grid topology or positional encoding, the model is fully inductive and can process cascade samples from any grid without retraining.
Training on cascade samples from distinct grids further regularizes the learned representations, preventing overfitting to any single topology.

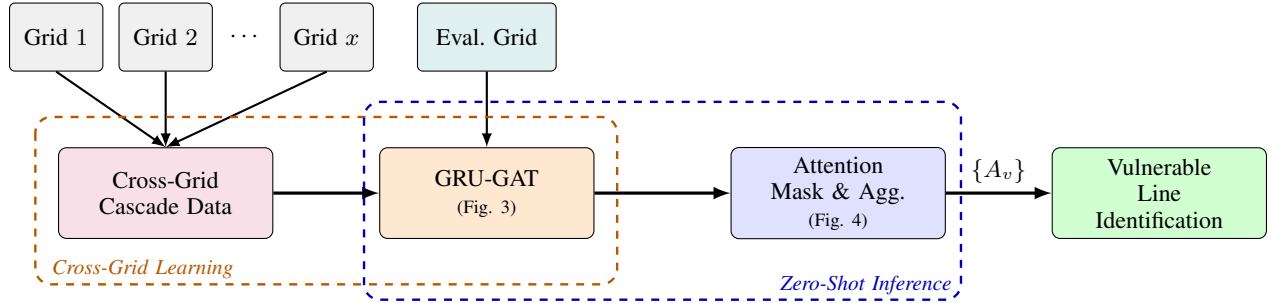
\begin{figure*}[t]
  \centering
  \begin{tikzpicture}[
    every node/.style={font=\small},
    phase/.style={rectangle, draw, rounded corners=3pt, minimum height=12mm,
                  text width=26mm, align=center},
    gridbox/.style={rectangle, draw, rounded corners=2pt, fill=gray!12,
                    minimum height=9mm, text width=10mm, align=center,
                    font=\small},
    evalbox/.style={rectangle, draw, rounded corners=2pt, fill=teal!12,
                    minimum height=9mm, text width=16mm, align=center,
                    font=\small},
    arr/.style={-latex, thick},
    bigarr/.style={-latex, line width=1.2pt},
    zonelabel/.style={font=\footnotesize\itshape},
  ]

  \node[phase, fill=purple!12] (comb) at (0,0)
        {Cross-Grid\\Cascade Data};

  \node[gridbox, above=10mm of comb] (g2) {Grid $2$};
  \node[gridbox, left=2mm of g2]  (g1) {Grid $1$};
  \node[right=1mm of g2, font=\small] (gdots) {$\cdots$};
  \node[gridbox, right=1mm of gdots] (gx) {Grid $x$};

  \draw[arr] (g1.south) -- (comb.north);
  \draw[arr] (g2.south) -- (comb.north);
  \draw[arr] (gx.south) -- (comb.north);

  \node[phase, fill=orange!18, right=14mm of comb] (model)
        {GRU-GAT\\{\scriptsize (Fig.~\ref{fig:grugat})}};

  \draw[bigarr] (comb) -- (model);

  \node[phase, fill=blue!12, right=18mm of model] (maskagg)
        {Attention\\Mask \& Agg.\\{\scriptsize (Fig.~\ref{fig:mask})}};

  \draw[bigarr] (model) -- (maskagg);

  \node[evalbox, fill=teal!12, above=10mm of model] (evalgrid)
        {Eval.\ Grid};
  \draw[arr] (evalgrid) -- (model.north);

  \node[phase, fill=green!18, right=14mm of maskagg] (rank)
        {Vulnerable\\Line\\Identification};

  \draw[bigarr] (maskagg) -- node[above, font=\small] {$\{A_v\}$} (rank);

  \draw[dashed, rounded corners=6pt, orange!70!black, line width=0.9pt]
    ($(comb.north west)+(-3mm,4mm)$) rectangle ($(model.south east)+(3mm,-6mm)$);
  \node[zonelabel, anchor=north west,text=orange!70!black]
    at ($(comb.south west)+(-2mm,-1.5mm)$)
    {Cross-Grid Learning};

  \draw[dashed, rounded corners=6pt, blue!70!black, line width=0.9pt]
    ($(model.north west)+(-2mm,6mm)$) rectangle ($(maskagg.south east)+(2mm,-8mm)$);
  \node[zonelabel, anchor=north east,text=blue!70!black]
    at ($(maskagg.south east)+(2mm,-3.5mm)$)
    {Zero-Shot Inference};

  \end{tikzpicture}
  \caption{CG-CAE pipeline overview.
  \emph{Cross-Grid Learning} (orange box): cascade data from multiple training grids are combined and used to train a single GRU-gated GAT (Fig.~\ref{fig:grugat}).
  \emph{Zero-Shot Inference} (blue box): at test time, cascade samples from an unseen evaluation grid are fed directly to the trained model --- no retraining --- and the resulting attention coefficients are filtered and aggregated (Fig.~\ref{fig:mask}).
  The GRU-GAT sits in the overlap of both phases: trained during cross-grid learning, then applied without modification during inference.
  The per-line cascade exposure scores $\{A_v\}$ (Eq.~\eqref{eq:agg}) rank all lines; higher scores indicate greater vulnerability to cascading failure propagation.}
  \label{fig:cae}
\end{figure*}

Fig.~\ref{fig:cae} illustrates the complete CG-CAE pipeline.
The GRU-GAT model is trained once on combined cascade data from the training grids; at inference time, evaluation cascade samples are fed directly to the trained model without any gradient updates or fine-tuning.

Three design properties of the CG-CAE pipeline are worth highlighting.
First, the pipeline follows a \emph{one model, many grids} paradigm: because all operations are inductive, a single model trained on combined data from the training grids applies directly to any evaluation grid, with no per-grid fine-tuning and no need to select which training grid best matches the target.
Second, \emph{cross-grid regularization} arises naturally from training on cascade samples drawn from distinct topologies; the GRU gate absorbs this diversity into a general update rule rather than memorizing grid-specific cascade patterns.
Third, the inference-time analysis is \emph{few-sample and label-free}: the pre-trained model runs in inference mode on cascade samples from the evaluation grid --- no gradient updates, no fine-tuning --- and the cascade exposure score emerges from the attention coefficients alone.
In practice, fewer than 100 cascade samples suffice for stable rankings, making the analysis efficient even for large grids.

\section{Experiments and Results}\label{Experiments}\label{Results}

\subsection{Dataset}
\label{Experiments:Dataset}

We use the PyPSA-EUR dataset, which provides geographically-referenced simulation and heuristic estimations of European power grids~\cite{PyPSAEur}.
Two snapshot years are used: 2013 grids serve as the training corpus, and 2019 grids serve as the evaluation corpus.
Table~\ref{tab:datasets} summarizes the three training grids and six evaluation grids.

\begin{table}[t]
  \centering
  \caption{PyPSA-EUR grids used for training and evaluation.}
  \label{tab:datasets}
  \begin{tabular}{llcc}
    \toprule
    Role   & Grid & Year & \#Lines \\
    \midrule
    Training & Portugal (PT)  & 2013 & 203  \\
    Training & Germany  (DE)  & 2013 & 433  \\
    Training & France   (FR)  & 2013 & 860  \\
    \midrule
    Evaluation & Portugal (PT)  & 2019 & 156  \\
    Evaluation & Switzerland (CH) & 2019 & 171  \\
    Evaluation & Great Britain (GB) & 2019 & 479 \\
    Evaluation & Germany  (DE)  & 2019 & 827  \\
    Evaluation & Spain     (ES) & 2019 & 832  \\
    Evaluation & France   (FR)  & 2019 & 1{,}459 \\
    \bottomrule
  \end{tabular}
\end{table}

Cascade samples are produced by the CF simulator described in Section~\ref{CAE:CF}.
For each training grid, we create a large pool of random initial-failure scenarios and select up to 30{,}000 samples using depth-weighted oversampling, which ensures that rare deep cascades are adequately represented.
Each node in a cascade sample carries its failure iteration as a class label: $g_u = 0$ for operational lines, $g_u = 1$ for initial failures, and $g_u \geq 2$ for lines that failed at cascade iteration $g_u$.
The cascade samples from the three training grids are concatenated into a single combined dataset for cross-grid training.

For evaluation, we produce a separate held-out pool of 1{,}000 cascade samples per evaluation grid.
These samples are excluded from both training and cascade exposure extraction, so they provide independent ground truth.
Table~\ref{tab:cascade_stats} summarizes the average cascade scale and depth across this held-out pool.
Cascade depth varies substantially across grids, ranging from an average of 2.6 iterations on CH to 16.0 on FR.
GB's disproportionately high scale and depth relative to its size may reflect the island grid's reliance on a small number of high-capacity corridors, which concentrate redistribution stress when failures occur.

\begin{table}[t]
  \centering
  \caption{Average cascade characteristics of the 1{,}000 held-out evaluation samples per grid.
           Scale is the mean percentage of lines failing beyond initial failures.
           Depth is the mean maximum cascade iteration.}
  \label{tab:cascade_stats}
  \begin{tabular}{lccc}
    \toprule
    Grid & \#Lines & Avg.\ Scale (\%) & Avg.\ Depth \\
    \midrule
    PT  & 156    & 2.8  & 3.5 \\
    CH  & 171    & 3.7  & 2.6 \\
    GB  & 479    & 18.1 & 8.9 \\
    DE  & 827    & 10.8 & 6.8 \\
    ES  & 832    & 4.5  & 5.4 \\
    FR  & 1{,}459 & 21.1 & 16.0 \\
    \bottomrule
  \end{tabular}
\end{table}

\subsection{Pre-Training Setup}
\label{Experiments:Pretrain}

 CG-CAE is trained on the combined cascade samples from the three training grids for up to 20 epochs with early stopping (patience 10).
We use the Adam optimizer with learning rate $5 \times 10^{-5}$, gradient accumulation over 4 steps, and cosine annealing with warm restarts.
The architecture uses hidden dimension $d{=}256$, four attention heads, and $C{=}100$ output classes.
The training objective is cross-entropy loss over the $C$ output classes.

\subsection{Zero-Shot Evaluation}
\label{Experiments:Zeroshot}
\label{Results:Zeroshot}

After training, we apply the model directly to each of the six evaluation grids described in Table~\ref{tab:datasets}, without any fine-tuning or retraining.
As noted above, the self-supervised formulation means the model receives each line's failure iteration as input and reconstructs the cascade pattern.
The classification task is therefore reconstructive, and high accuracy is expected.
 CG-CAE achieves macro-F1 of 1.00 on all three training grids and on all six evaluation grids without any fine-tuning.
The three inter-time grids (PT, DE, FR at 2019) have different network sizes from the training data, and the three inter-domain grids (CH, GB, ES) share no lines with the training corpus.
The model's practical value lies not in this classification accuracy but in the attention weights it produces during inference, which are extracted for cascade exposure analysis below.
The perfect macro-F1 also indicates that the failure-iteration embedding provides a sufficient input representation, leaving no room for additional physical features to further benefit the model.

\subsection{Cascade Exposure}
\label{Experiments:Ranking}
\label{Results:Ranking}

The following four subsections evaluate cascade exposure from complementary angles.

\subsubsection{Baselines}
\label{Experiments:Baselines}
We choose EB and PR as baselines.
Both rely on the grid's topological and electrical properties and do not learn from cascade data.
We are not aware of any existing GNN method that performs zero-shot line-level vulnerability  identification, so a learned-baseline comparison is not currently available.
\begin{itemize}
  \item \textit{Electric Betweenness (EB)} --- measures the fraction of power flow passing through each line under a uniform injection scenario~\cite{EB23Chen}.
  \item \textit{PageRank (PR)} --- the stationary distribution of a modified PageRank, weighted by the Branch Outage Distribution Factor (BODF) matrix, computed directly from the power network~\cite{pagerank23fan}
\end{itemize}

\subsubsection{Mean Top-$\tau\%$ Vulnerability}
\label{Experiments:MeanVul}
\label{Results:MeanVul}

\begin{figure*}[t]
  \centering
  \includegraphics[width=\textwidth]{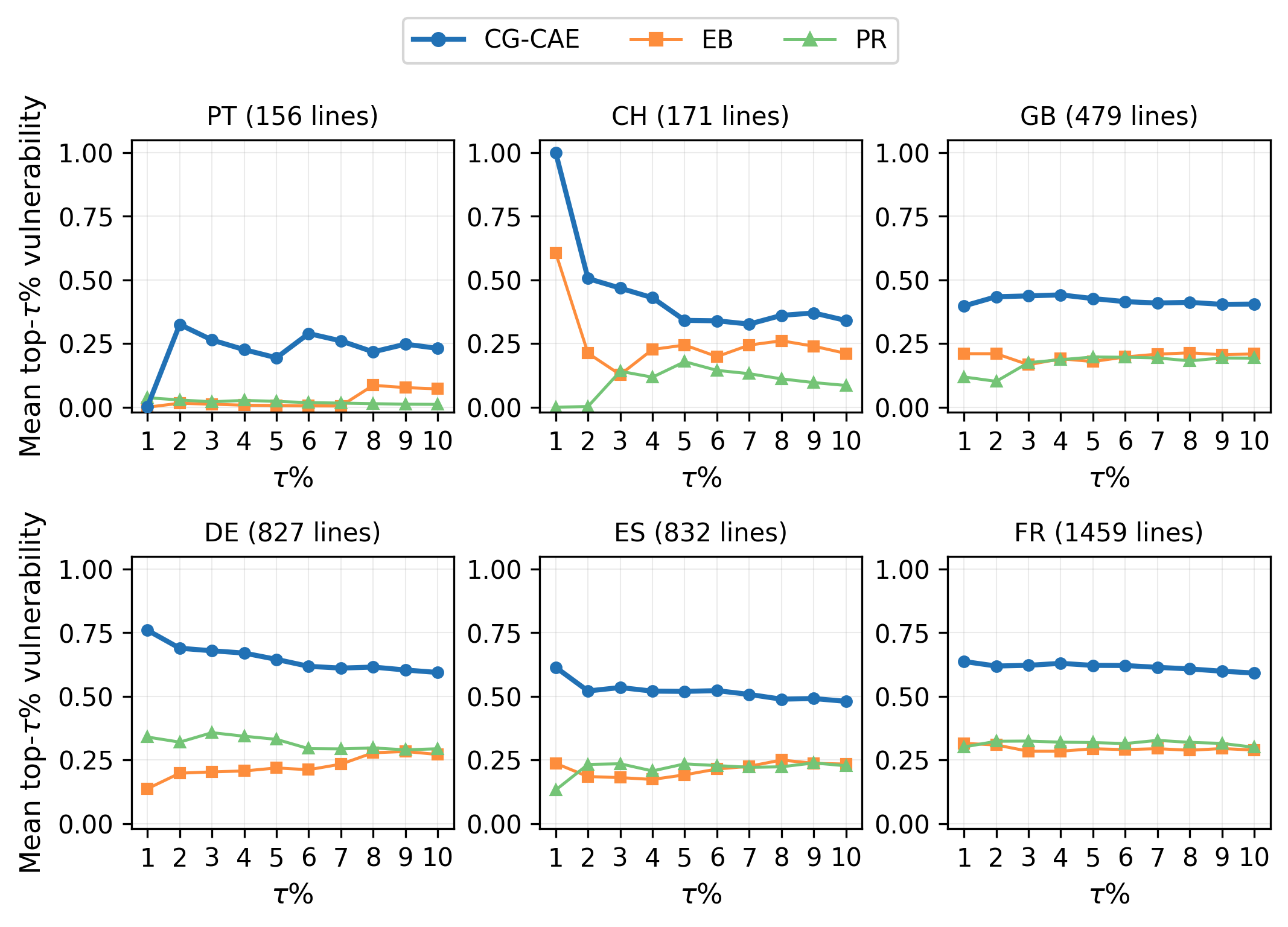}
  \caption{Mean top-$\tau$\% vulnerability on six evaluation grids (100 cascade samples).
           Each panel shows one grid; lines represent  CG-CAE, EB, and PR.
            CG-CAE (blue) sits above both baselines at every grid and threshold.}
  \label{fig:cascade_exposure_lines}
\end{figure*}

For a given ranking, we select the top $\tau\%$ of lines by their exposure score and compute the mean ground-truth vulnerability of those lines.
We sweep $\tau \in \{1, 2, \ldots, 10\}\%$ for all six evaluation grids.
Figure~\ref{fig:cascade_exposure_lines} shows the mean top-$\tau\%$ vulnerability for  CG-CAE, EB, and PR.
 CG-CAE's ranking is derived from 100 cascade samples, while EB and PR are computed from the grid's topological and electrical properties.
Ground-truth vulnerability is evaluated against the 1{,}000 held-out cascades.
Each panel corresponds to one evaluation grid, with $\tau$ on the horizontal axis and mean vulnerability on the vertical axis.
Higher curves indicate that the method's top-ranked lines are more vulnerable.

 CG-CAE sits above both baselines on every grid and at every threshold.
Averaged across grids at top-10\%,  CG-CAE achieves a mean vulnerability of 0.441, compared with 0.215 for EB and 0.185 for PR.
The gap is widest on DE (0.594 vs.\ 0.294) and narrowest on PT, where the small grid size (156 lines) means that 1\% corresponds to only one or two lines.

\subsubsection{Sample Efficiency}
\label{Experiments:Efficiency}
\label{Results:Efficiency}

\begin{figure*}[t]
  \centering
  \includegraphics[width=\textwidth]{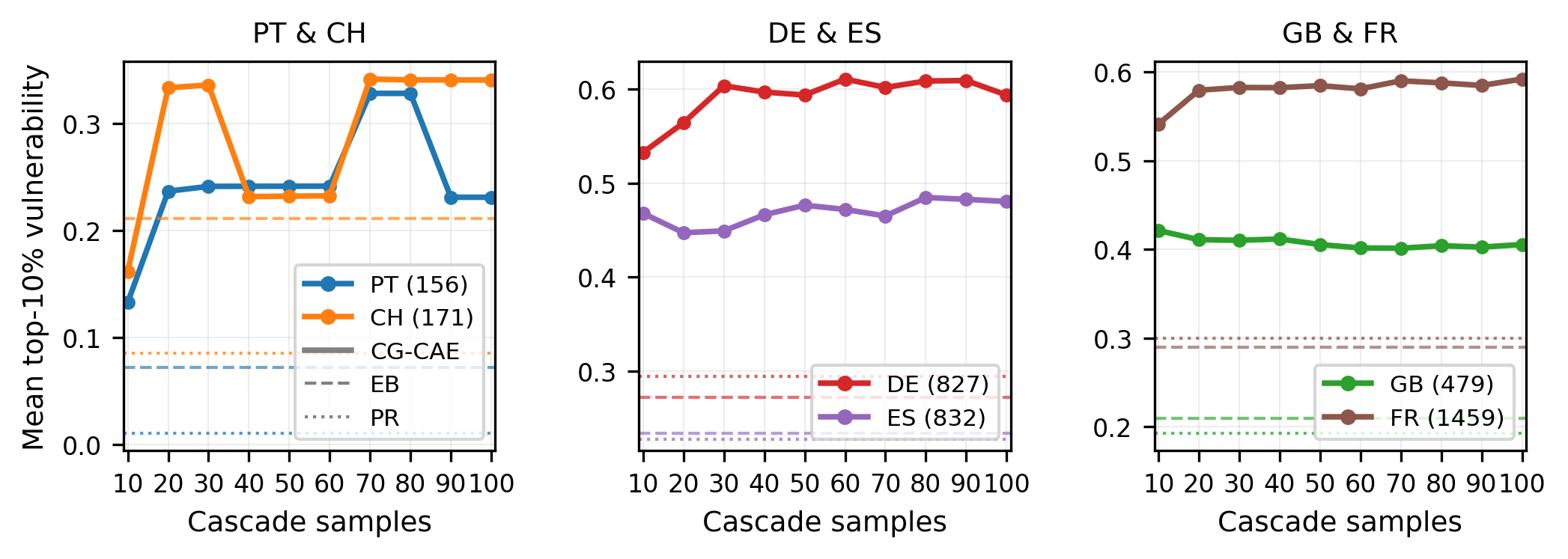}
  \caption{Per-grid sample efficiency of cascade exposure (top-10\%).
           Grids are grouped by cascade scale and depth (Table~\ref{tab:cascade_stats}).
           Solid lines show CG-CAE; dashed and dotted horizontal lines show per-grid EB and PR baselines.}
  \label{fig:efficiency}
\end{figure*}

To assess how quickly cascade exposure stabilizes, we sweep the number of cascade samples $N_s$ from 10 to 100 in increments of 10.
Figure~\ref{fig:efficiency} shows the mean top-10\% vulnerability for each evaluation grid as $N_s$ increases, with grids grouped by cascade scale and depth (Table~\ref{tab:cascade_stats}).
All six grids exceed their respective EB and PR baselines at every sample count.
Grids with higher cascade scale and depth (GB, FR) converge smoothly, consistent with the richer propagation signal each sample provides.
PT and CH, whose cascades are shallow and limited, show more volatility at low $N_s$.
Rankings stabilize by approximately 30 samples across all grids.

\subsubsection{Cascade Exposure Score Evaluation}
\label{Experiments:ExposureScore}
\label{Results:ExposureScore}

Mean top-$\tau\%$ vulnerability evaluates how vulnerable the selected lines are, but it does not reveal whether the method surfaces all genuinely vulnerable lines.
We therefore evaluate the mean percentile rank of the high-exposure set, a threshold-free metric that measures how well the cascade exposure score $A_v$ identifies lines with high ground-truth vulnerability.
We define a line as having high exposure if its ground-truth vulnerability exceeds the median vulnerability among all lines that failed in at least one held-out cascade.
Lines that never failed in any held-out cascade are excluded from this median calculation, so the threshold reflects the midpoint of the observed failure-prone population.
Let $\mathcal{E}$ denote the resulting high-exposure set and $L$ the total number of lines in the grid.
For a given method, let $r(v) \in \{1, \ldots, L\}$ be the rank assigned to line $v$ (1 = highest score).
The mean percentile rank is:
\begin{equation}
  \text{MPR} = \frac{1}{|\mathcal{E}|} \sum_{v \in \mathcal{E}} \frac{r(v)}{L}
  \label{eq:mpr}
\end{equation}
A perfect ranker that places every high-exposure line at the top achieves $\text{MPR} \approx |\mathcal{E}|/(2L)$, because the mean rank of the top $|\mathcal{E}|$ positions is $(|\mathcal{E}|+1)/2$.
A random ranker achieves $\text{MPR} = 0.5$ in expectation.
Lower values indicate that the method concentrates high-exposure lines near the top of its ranking.

\begin{figure}[t]
  \centering
  \includegraphics[width=\columnwidth]{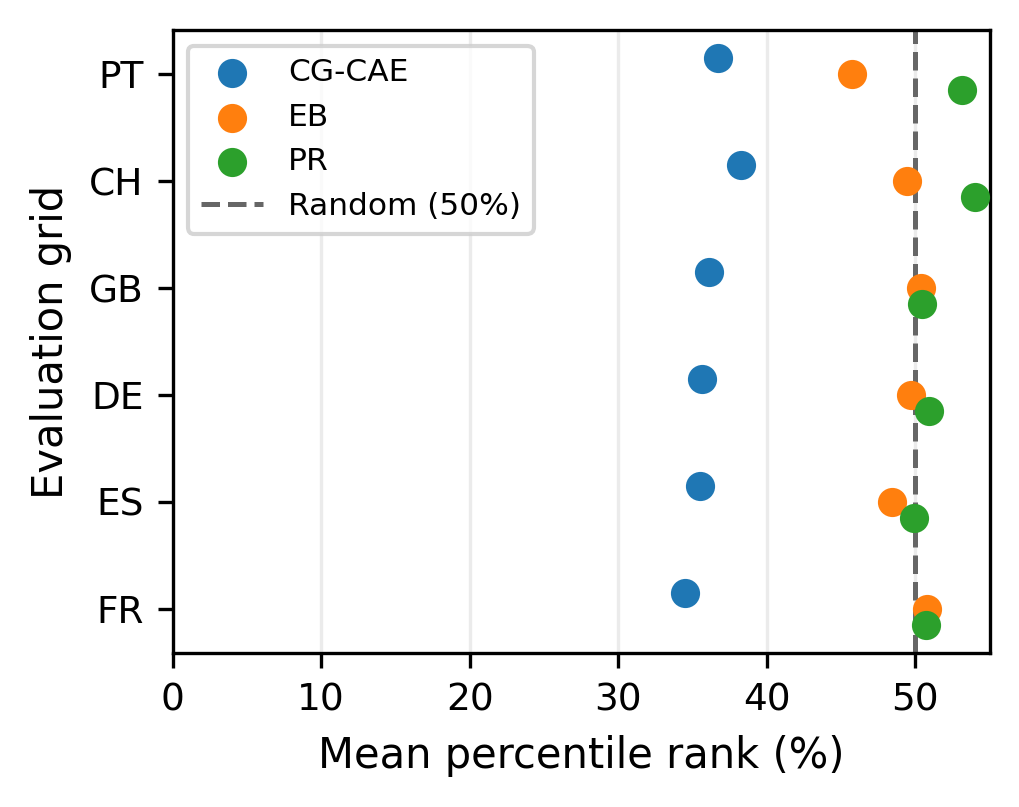}
  \caption{Mean percentile rank of the high-exposure set under each method's cascade exposure score.
           Lower is better; the dashed line at 50\% marks the random-ranker baseline.
            CG-CAE consistently places high-exposure lines closer to the top of its ranking.}
  \label{fig:exposure_score}
\end{figure}

Figure~\ref{fig:exposure_score} shows the mean percentile rank (Eq.~\ref{eq:mpr}) of the high-exposure set for each method and grid.
A random ranker would place these lines at the 50th percentile on average.
 CG-CAE achieves a mean percentile rank of 36.1\% averaged across six grids, compared with 49.1\% for EB and 51.5\% for PR.
On every evaluation grid,  CG-CAE places the high-exposure population closer to the top of its ranking than either baseline.

\subsection{Depth-Stratified Analysis}
\label{Experiments:DepthAnalysis}
\label{Results:DepthAnalysis}

The aggregate cascade exposure results above do not reveal whether  CG-CAE's advantage is uniform across depths or concentrated at specific cascade iterations.
We partition lines into two depth groups using the floor of the average cascade depth $\lfloor \bar{d} \rfloor$ from Table~\ref{tab:cascade_stats}.
The shallow group ($d_u \leq \lfloor \bar{d} \rfloor$) contains lines that fail within the grid's typical cascade reach.
The deep group ($d_u > \lfloor \bar{d} \rfloor$) contains lines that fail beyond the typical cascade depth and require longer propagation chains.
Using a per-grid cutoff ensures that each grid's depth groups reflect its own cascade-depth distribution.
For example, CH has an average depth of 2.6, so $\lfloor \bar{d} \rfloor = 2$ and only iteration-2 failures fall in the shallow group.
FR has an average depth of 16.0, so $\lfloor \bar{d} \rfloor = 16$ and only the tail of very deep cascades enters the deep group.
We evaluate both a precision angle and a recall angle.
For the precision angle, we compute depth-specific ground-truth vulnerability from held-out cascades and report the mean shallow and deep vulnerability of each method's global top-$\tau\%$ lines.
For the recall angle, we define a depth-conditional high-exposure set in each group and report its mean percentile rank.

\subsubsection{Precision Angle}
\label{Results:DepthVul}

\begin{figure*}[t]
  \centering
  \includegraphics[width=\textwidth]{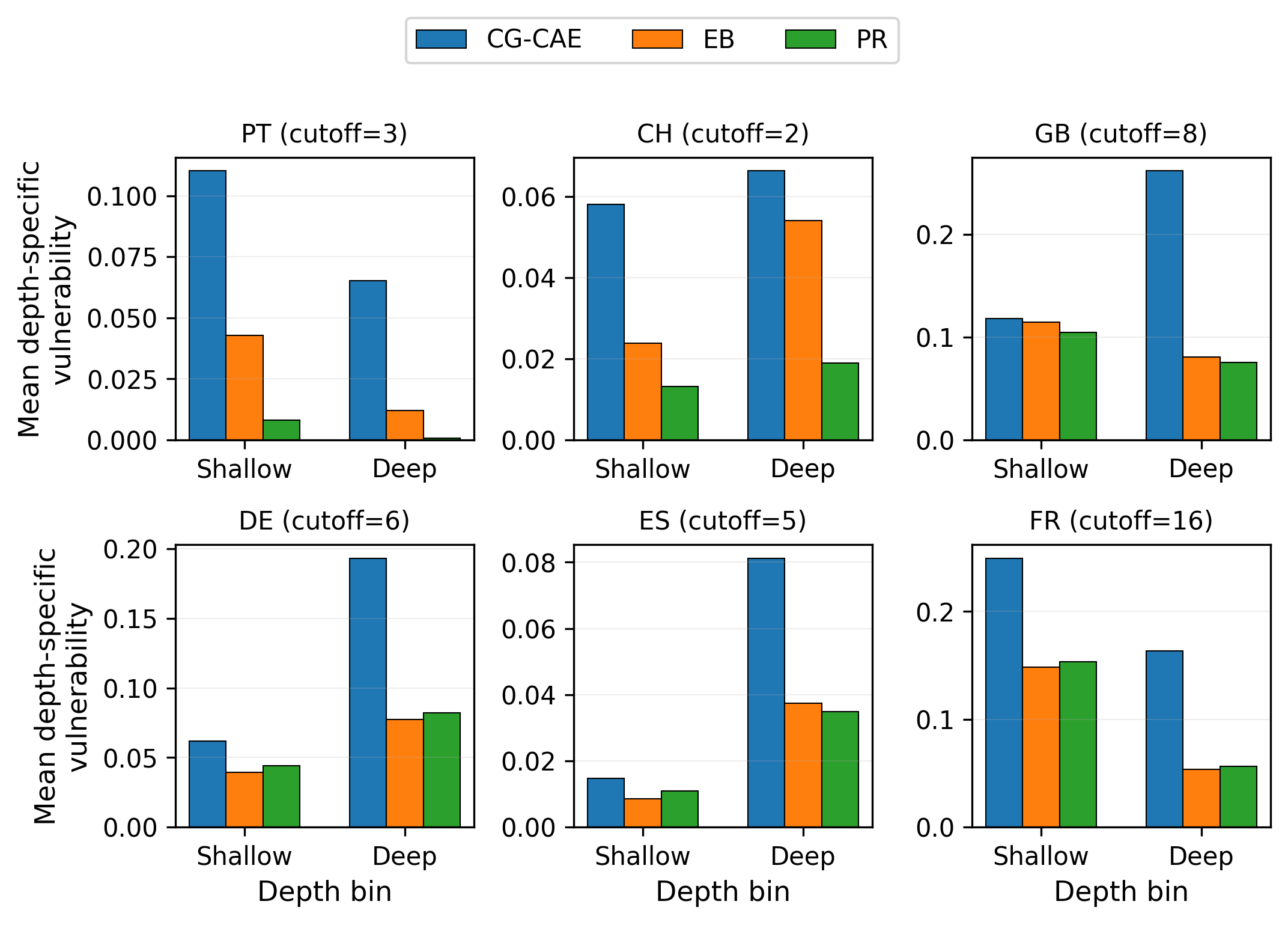}
  \caption{Depth-stratified mean top-10\% vulnerability (precision angle) per evaluation grid.
           Each grid uses a per-grid cutoff $\lfloor \bar{d} \rfloor$.
           Bars show the mean ground-truth vulnerability of each method's global top-10\% lines that fall in each bin.}
  \label{fig:depth_vul}
\end{figure*}

For each line, we compute shallow and deep ground-truth vulnerability from the held-out cascades based on whether failures occur at iterations $\leq \lfloor \bar{d} \rfloor$ or $> \lfloor \bar{d} \rfloor$.
Table~\ref{tab:depth_2bin} reports, for each method's global top-10\% lines, the mean shallow and deep vulnerability.
Figure~\ref{fig:depth_vul} shows the same pattern per grid.

\begin{table}[t]
  \centering
  \caption{Mean depth-specific vulnerability of each method's global top-10\% lines.
            CG-CAE's ranking uses 100 cascade samples; vulnerability values are computed against the 1{,}000 held-out cascades.
           \textbf{Bold} marks the best method per grid and depth group.}
  \label{tab:depth_2bin}
  \setlength{\tabcolsep}{3pt}
  \footnotesize
  \begin{tabular}{llcccccc}
    \toprule
    Bin & Method & PT & CH & GB & DE & ES & FR \\
    \midrule
    \multirow{3}{*}{Shallow}
    &  CG-CAE  & \textbf{0.110} & \textbf{0.058} & \textbf{0.118} & \textbf{0.062} & \textbf{0.015} & \textbf{0.250} \\
    & EB      & 0.043 & 0.024 & 0.115 & 0.039 & 0.009 & 0.148 \\
    & PR      & 0.008 & 0.013 & 0.105 & 0.044 & 0.011 & 0.153 \\
    \midrule
    \multirow{3}{*}{Deep}
    &  CG-CAE  & \textbf{0.065} & \textbf{0.066} & \textbf{0.262} & \textbf{0.193} & \textbf{0.081} & \textbf{0.163} \\
    & EB      & 0.012 & 0.054 & 0.081 & 0.077 & 0.037 & 0.054 \\
    & PR      & 0.001 & 0.019 & 0.075 & 0.082 & 0.035 & 0.056 \\
    \bottomrule
  \end{tabular}
\end{table}

Across all six grids and both bins,  CG-CAE yields the highest mean depth-specific vulnerability for its global top-10\% lines.
The deep-bin advantage is especially pronounced on GB (0.262 vs. 0.081/0.075) and DE (0.193 vs. 0.077/0.082), where cascade propagation chains are longer.
The same pattern appears on ES and FR, while PT and CH also show consistent gains despite their smaller sizes.
These precision-side results align with Figure~\ref{fig:depth_rank}, where deep high-exposure lines are ranked earlier by  CG-CAE.

\subsubsection{Recall Angle}
\label{Results:DepthRank}

\begin{figure*}[t]
  \centering
  \includegraphics[width=\textwidth]{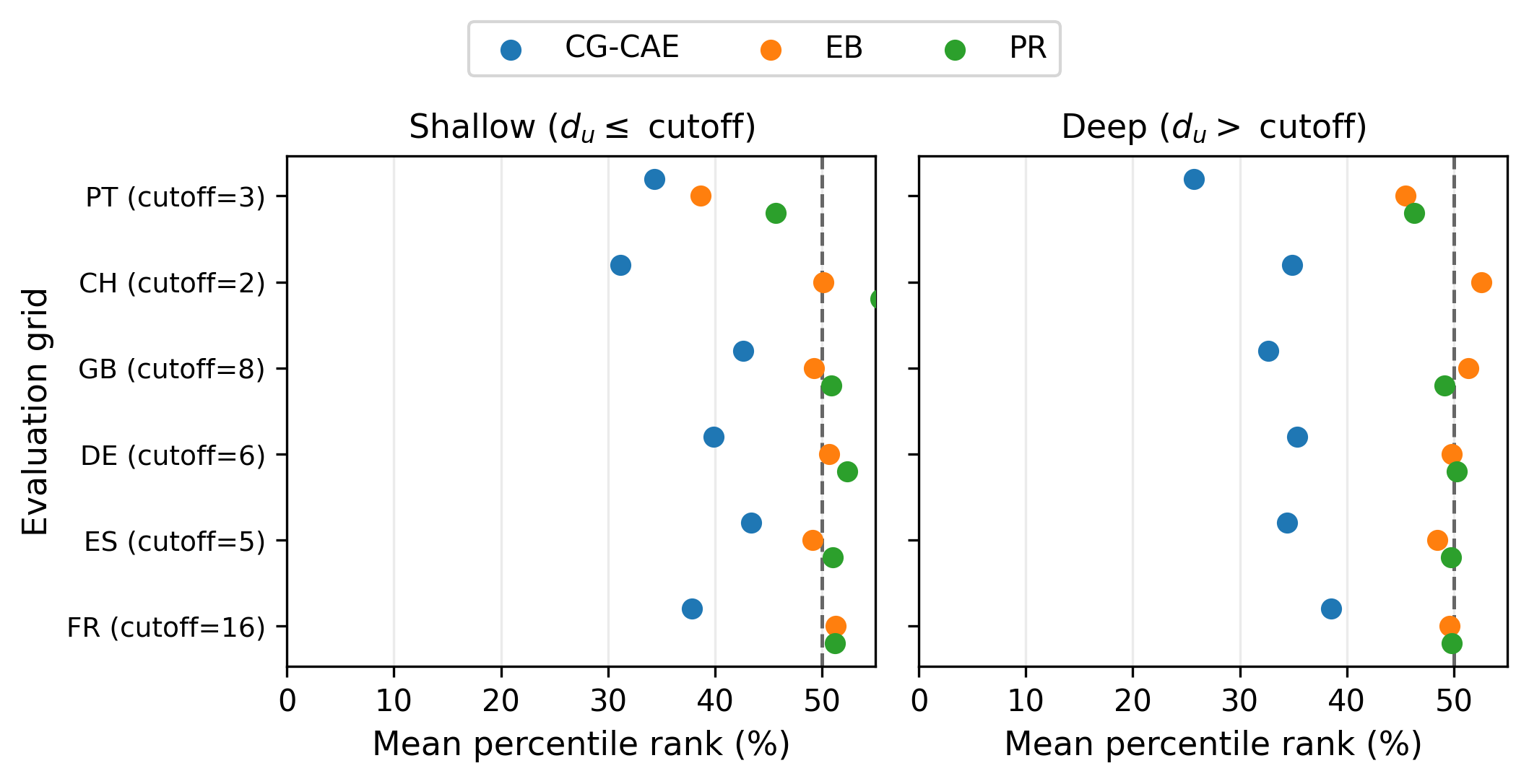}
  \caption{Mean percentile rank of the depth-conditional high-exposure set (recall angle), split into shallow ($d_u \leq \lfloor \bar{d} \rfloor$) and deep ($d_u > \lfloor \bar{d} \rfloor$) bins with per-grid cutoff $\lfloor \bar{d} \rfloor$.
           Lower is better; the dashed line at 50\% marks the random baseline.
           In the deep panel,  CG-CAE places high-exposure lines much closer to the top on every grid.}
  \label{fig:depth_rank}
\end{figure*}

Figure~\ref{fig:depth_rank} shows the mean percentile rank of the depth-conditional high-exposure set within each bin.
In the deep panel,  CG-CAE's points consistently shift far left of the baselines on every grid, confirming earlier retrieval of deep-cascade high-exposure lines.
In the shallow panel,  CG-CAE still leads on most grids, though the gap narrows compared with the deep panel.

\subsection{Discussion}
\label{Results:Discussion}

\subsubsection{Classification Accuracy Is Expected}
The perfect macro-F1 across all grids follows directly from the self-supervised formulation: the model receives each line's failure iteration as input and reconstructs the cascade pattern.
Because the task is reconstructive rather than predictive, high accuracy is not surprising.
The result confirms that the GRU-gated architecture retains grid-agnostic cascade logic that transfers without domain adaptation, but the model's practical value lies in the attention weights we extract, not in the classification score itself.

\subsubsection{CG-CAE Identifies More Vulnerable Lines at Every Threshold}
Figure~\ref{fig:cascade_exposure_lines} shows that  CG-CAE's top-ranked lines are consistently more vulnerable than those selected by EB or PR, across all six grids and all $\tau$ values.
The gap is widest on larger grids such as DE and FR, where longer propagation chains create a richer signal for the cascade-depth mask to exploit.
On PT, the gap narrows because the small grid size (156 lines) limits the diversity of cascade paths available.

\subsubsection{Cascade Exposure Is Sample-Efficient}
As Figure~\ref{fig:efficiency} shows, CG-CAE exceeds both baselines on every grid with as few as 10 cascade samples and stabilizes by approximately 30.
This is practically significant because cascade simulation is computationally expensive, and requiring fewer samples reduces the cost of deploying CG-CAE on a new grid.
PT and CH show greater volatility at low $N_s$ because their shallow cascade depth limits the propagation signal available per sample.

\subsubsection{Small Grids Are Noisy at Narrow Thresholds}
PT (156 lines) and CH (171 lines) are the two smallest evaluation grids.
At top-1\%, the ranking metric is sensitive to individual line placements.
PT recovers to 0.194 at top-5\% and 0.231 at top-10\%.
CH's top-1\% cascade exposure of 1.000 reflects a single highly vulnerable line being selected, and this result should be interpreted cautiously.
On CH, EB also scores well (0.607 at top-1\%), possibly because a compact grid topology concentrates power flow on a small number of heavily loaded lines that EB is designed to capture.

\subsubsection{Cascade Exposure Score Confirms the Recall-Side Advantage}
Figure~\ref{fig:exposure_score} shows that  CG-CAE not only assigns high scores to vulnerable lines (precision, Figure~\ref{fig:cascade_exposure_lines}) but also concentrates the full high-exposure population near the top of its ranking.
A grid operator using  CG-CAE for prioritized inspection would encounter high-exposure lines earlier in the ranked list than with either baseline.

\subsubsection{The Advantage Is Concentrated in Deep Cascades}
Figure~\ref{fig:depth_vul} and Table~\ref{tab:depth_2bin} show that  CG-CAE's mean vulnerability advantage over the best baseline is generally larger in the deep group ($d_u > \lfloor \bar{d} \rfloor$) than in the shallow group ($d_u \leq \lfloor \bar{d} \rfloor$).
The depth-stratified rank analysis (Figure~\ref{fig:depth_rank}) corroborates this from the recall side:  CG-CAE's mean percentile rank is consistently lower (better) than both baselines in the deep panel.
This pattern arises because the cascade-depth mask progressively opens at later GRU-GAT steps, admitting source nodes that have accumulated multi-hop cascade context.
The result is a richer attention signal for lines that fail through extended propagation chains -- precisely where static metrics lack information.
At shallow depths, fewer GRU-GAT steps have executed and the mask admits fewer informed source nodes, so  CG-CAE's propagation advantage is inherently sparser.
The contrast between depth groups confirms that  CG-CAE's advantage stems from modeling the cascading failure propagation mechanism, not from static grid properties.

\subsubsection{The Advantage Over Static Baselines Is Consistent}
EB and PR capture structural and electrical importance but cannot model multi-step propagation dynamics.
 CG-CAE's consistent advantage across all grids and thresholds confirms that the learned cascade correlations carry information beyond what static grid properties provide.

\section{Conclusion}\label{Conclusion}

We presented  CG-CAE, a cross-grid self-supervised framework for  inductive power grid cascading failure analysis.
The framework  rests on three design elements: a GRU-gated graph attention mechanism that propagates cascade state recurrently; a cascade-depth mask that filters attention to causally informed source nodes; and cross-grid training that exposes a single model to diverse topologies.

A single model trained on three European grids transfers to six structurally diverse evaluation grids without fine-tuning.
The cascade exposure score extracted from the trained attention weights identifies more vulnerable lines than Electric Betweenness and PageRank at every tested top-$\tau$\% threshold, with a mean top-10\% vulnerability of 0.441 versus 0.215 and 0.185 for the two baselines.
The depth-stratified analysis reveals where this advantage originates:  CG-CAE's gain over the baselines is larger in the deep-cascade group ($d_u > \lfloor \bar{d} \rfloor$) than in the shallow group, particularly on grids with extended propagation chains such as GB and DE.
This pattern confirms that the cascade-depth mask and recurrent propagation together capture multi-step failure correlations that static topology metrics cannot access.
The inference-time analysis is also sample-efficient, stabilizing with approximately 30 cascade samples per grid.

\section*{Acknowledgment}
Xiang Li is supported by NSF CNS-1948550.

The authors used Claude (Anthropic) and GitHub Copilot (Microsoft) to assist with code debugging and prose refinement.

\bibliographystyle{IEEEtran}
\bibliography{reference}
\end{document}